\pdfoutput=1

\documentclass[11pt]{article}

\usepackage{acl}

\usepackage{times}
\usepackage{latexsym}
\usepackage{multirow}   
\usepackage{graphicx}   
\usepackage{enumitem}
\usepackage{amsmath}    
\usepackage{amssymb}    
\usepackage{latexsym}
\usepackage[inkscapelatex=false]{svg}
\usepackage{url}    
\usepackage{colortbl, booktabs} 
\usepackage{makecell} 
\usepackage{color}  
\usepackage{array}
\usepackage{hyperref}   
\usepackage[all]{nowidow}   
\usepackage[subtle]{savetrees} 

\usepackage{algorithm}
\usepackage{algorithmic}
\usepackage{enumitem}
\usepackage{microtype}
\usepackage{multicol}
\usepackage{tipa}
\usepackage{arydshln} 
\usepackage{tipa}

\usepackage{inconsolata}

\usepackage[T1]{fontenc}

\usepackage[utf8]{inputenc}

\title{\textit{Comparing~Apples~to~Oranges}:\\ A~Dataset~\&~Analysis~of~LLM~Humour~Understanding from Traditional Puns to Topical Jokes}

\author{Tyler Loakman\textsuperscript{1},
William Thorne\textsuperscript{1},
Chenghua Lin\textsuperscript{2}\\
  \textsuperscript{1}Department of Computer Science, The University of Sheffield, UK \\
  \textsuperscript{2}Department of Computer Science, The University of Manchester, UK\\
  \texttt{tcloakman1@sheffield.ac.uk} \\
  \texttt{wthorne1@sheffield.ac.uk} \\
  \texttt{chenghua.lin@manchester.ac.uk}}

\begin{document}

\maketitle

\begin{abstract}
Humour, as a complex language form, is derived from myriad aspects of life. Whilst existing work on computational humour has focussed almost exclusively on short pun-based jokes, we investigate whether the ability of Large Language Models (LLMs) to explain humour depends on the particular form. We compare models' joke explanation abilities from simple puns to complex topical humour that requires esoteric knowledge of real-world entities and events. To this end, we curate a dataset of 600 jokes across 4 joke types and manually write high-quality explanations. These jokes include heterographic and homographic puns, contemporary internet humour, and topical jokes. Using this dataset, we compare the zero-shot abilities of a range of LLMs to accurately and comprehensively explain jokes of different types, identifying key research gaps in the task of humour explanation. We find that none of the tested models (including reasoning models) are capable of reliably generating adequate explanations of all joke types, further highlighting the narrow focus of most existing works on overly simple joke forms.\footnote{\textcolor{red}{\textbf{CONTENT WARNING:} Some of the presented jokes may be considered offensive or distasteful to different individuals.}}

\end{abstract}

\section{Introduction}
\label{sec:intro}

The perception and understanding of humour is essential to human interaction and entertainment. Consequently, the ability to correctly process jokes is paramount to achieving a human-like understanding of ambiguity and social context in language. However, most existing work on computational humour has narrowly focussed on the simpler task of humour detection \cite{loakman-etal-2023-iron, meaney-etal-2021-semeval, miller-etal-2017-semeval}, which allows models to learn from confounding signals to detect certain joke types \cite{inacio-etal-2023-humor}. On the other hand, humour \textit{explanation} tasks a computational model with the more challenging objective of explicitly outlining how and why a presented text is humorous  \cite{hessel-etal-2023-androids}.

In addition to the general focus on joke detection, existing works on all humour processing tasks (i.e., detection, explanation, and generation) have mostly focussed on short-form puns such as "\textit{Yesterday I accidentally swallowed some food colouring. The doctor says I'm OK, but I feel like I've dyed a little inside}" \cite{he-etal-2019-pun}, where humour arises from the interpretation of the punning word (i.e., "dyed") with a phonetically similar word with different meaning (i.e., "died") in the case of heterographic puns, or the polysemy of a word for homographic puns \cite{Attardo+2008+101+156}. Whilst puns are commonplace, the humour is primarily self-contained, requiring only a common-sense understanding of the world (e.g., that food colouring "dyes" things) and an understanding of the semantics of dictionary words (e.g., dye/die).
On the other hand, much of the humour encountered online, on TV panel shows, and in stand-up comedy routines, is based on contemporary topical knowledge of evolving pop-culture phenomena and news events, where a full appreciation of the humour relies on potentially esoteric knowledge, rather than common-sense \cite{IJoC3611, Laineste_2002}, requiring highly complex reasoning and knowledge retrieval.

\begin{figure*}
    \centering
    \includegraphics[width=0.8\linewidth]{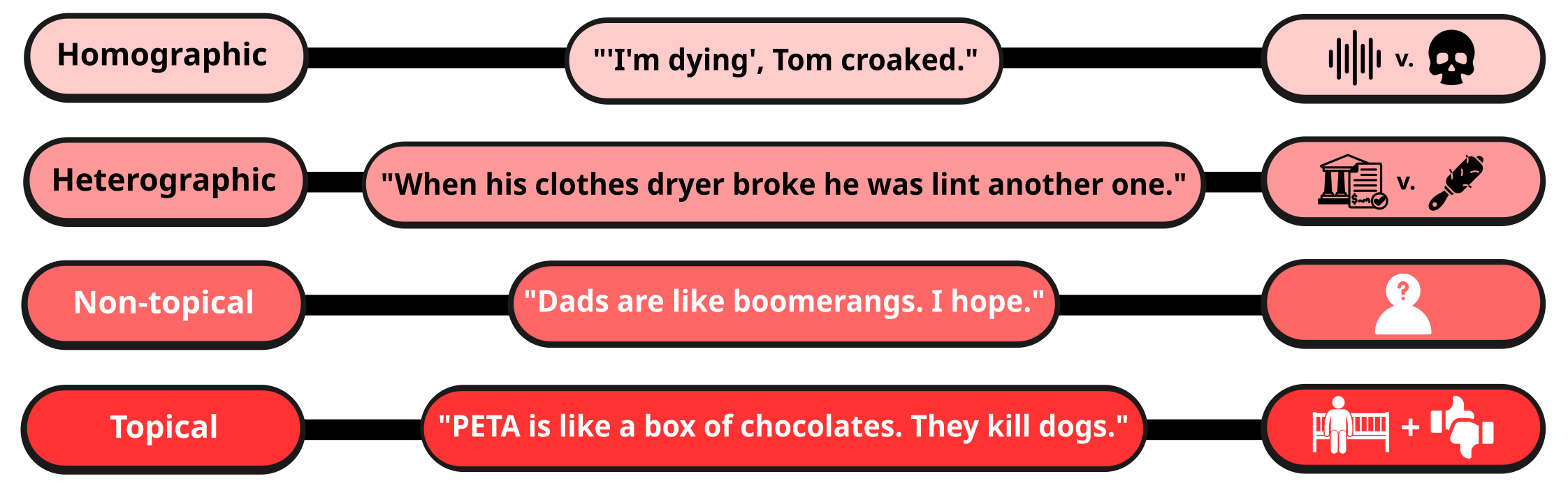}
    \caption{Examples of the 4 broad categories of joke compared in this work.}
    \label{fig:examples}
\end{figure*}

We present examples of each joke type that we investigate in \autoref{fig:examples}. The \textit{homographic} pun exploits the dual meaning of "croaked" as both a style of speech and a euphemism for dying; the \textit{heterographic} pun relies on the phonetic similarity between "lint" and "leant"; the \textit{non-topical} joke plays on the trope of an absentee father (not) returning like a boomerang; and the \textit{topical} joke refers to the animal welfare organisation PETA's high euthanasia rates and a reference to the movie Forrest Gump. Each joke type plays on polysemy, phonetics, social constructs, and esoteric knowledge, respectively.

To address this research gap, we assess the ability of Large Language Models (LLMs) to explain this range of joke forms in a zero-shot setting. As a result, we present the first work on assessing whether state-of-the-art LLMs are equally able to explain different joke formats, from those that require an understanding of basic semantics and phonetics, to those that require esoteric knowledge of pop culture and news events, therefore assessing the models' abilities to perform complex reasoning and knowledge retrieval.

We summarise our main contributions below. In presenting these contributions, we address the primary research questions of (\textbf{RQ1}) whether different joke formats have an impact on the ability of LLMs to explain humour and by extension, (\textbf{RQ2}) whether findings from existing work on computational humour are generalisable to the different jokes found in everyday life.

\begin{itemize} 
    \item We present a novel, balanced dataset of 600 jokes, categorised into puns, non-topical Reddit jokes, and topical Reddit jokes. Each joke is paired with a high-quality, succinct, human-authored reference explanation, enabling the first empirical assessment of how joke format affects the ability of LLMs to explain humour.\footnote{We release our dataset and code at: \url{https://github.com/tylerL404/Comparing-Apples-to-Oranges}.}
    \item We use the aforementioned dataset to assess the accuracy and completeness of 4800 joke explanations across 8 state-of-the-art open- and closed-source LLMs of various sizes, including reasoning models (600 jokes * 8 models). We analyse these explanations using human evaluation and automatic metrics, including using an LLM judge.
    \item We qualitatively evaluate the models' abilities to understand complex humour through LLM-generated explanations of a topical online joke in the form of a case study.
\end{itemize}

\section{Related Works}

\paragraph{Humour Generation}
Early humour generation efforts relied on simple template-based systems to create puns \cite{valitutti-etal-2013-everything, ritchie-2005-computational}, whilst contemporary works fine-tune deep learning models on extensive datasets of existing jokes \cite{garimella-etal-2020-judge,weller-etal-2020-humor,he-etal-2019-pun,petrovic-matthews-2013-unsupervised}. More recently, using different prompting strategies with LLMs has become the primary paradigm for generating jokes \cite{chen-etal-2024-u,mittal-etal-2022-ambipun}. Furthermore, \citet{sun-etal-2022-context} presented CUP, which turned the focus away from standalone joke generation to context-situated generation. Additionally, work has also been performed on humour-adjacent language forms such as tongue-twisters \cite{train_and_constrain,loakman-etal-2023-twistlist, keh-etal-2023-pancetta}. 

\paragraph{Humour Detection}
Similarly, humour detection has evolved from using lexical and syntactic feature-based methods \cite{van-den-beukel-aroyo-2018-homonym} to training language-model-based classifiers \cite{meaney-etal-2021-semeval,wang-etal-2020-unified, weller-seppi-2019-humor}. \citet{ao-etal-2022-combining} additionally present work using humour to aid in the detection of political parody, whilst \citet{meaney-2020-crossing} and \citet{loakman-etal-2023-iron} investigated the impact of demographic variables on humour perception. Furthermore, some works have approached humour detection from a multimodal standpoint, using elements such as audience laughter to aid in detection \cite{hasan-etal-2019-ur}.

\paragraph{Humour Explanation}
Lastly, humour explanation provides an account for why a given text is funny. Consequently, this task requires models to exhibit extensive reasoning capabilities. Whilst explaining standalone puns usually only requires a model to identify the pun word and its alternative interpretations in the given context, more complex jokes require an understanding of common sense, social constructs, and esoteric knowledge to identify key incongruities. Unlike the task of detection, explanation requires models to not only be able to identify humour, but also "understand" it to provide a textual explanation. \citet{miller-etal-2017-semeval} presented a shared task concerning humour, in which subtask 3 pertains to humour interpretation by tasking models with assigning the correct WordNet sense keys to given instances of pun words. \citet{inacio-etal-2023-humor} tackle the humour explanation task indirectly by investigating what specific elements humour classifiers learn. Regarding the production of natural language explanations, \citet{hessel-etal-2023-androids} perform the task of humour explanation on multimodal data from caption contests in The New Yorker, generating explanations as to why a particular best-rated caption is funny. Finally, \citet{drivelology} investigate the ability of LLMs to explain a wide range of linguistic phenomena that require non-linear reasoning, such as tautologies and sarcasm.

Our work empirically investigates the performance gap for joke explanation on different types of humour, assessing the challenges presented by contemporary topical jokes in particular.

\section{Dataset Compilation}

\begin{figure}
    \centering
    \includegraphics[width=0.80\linewidth]{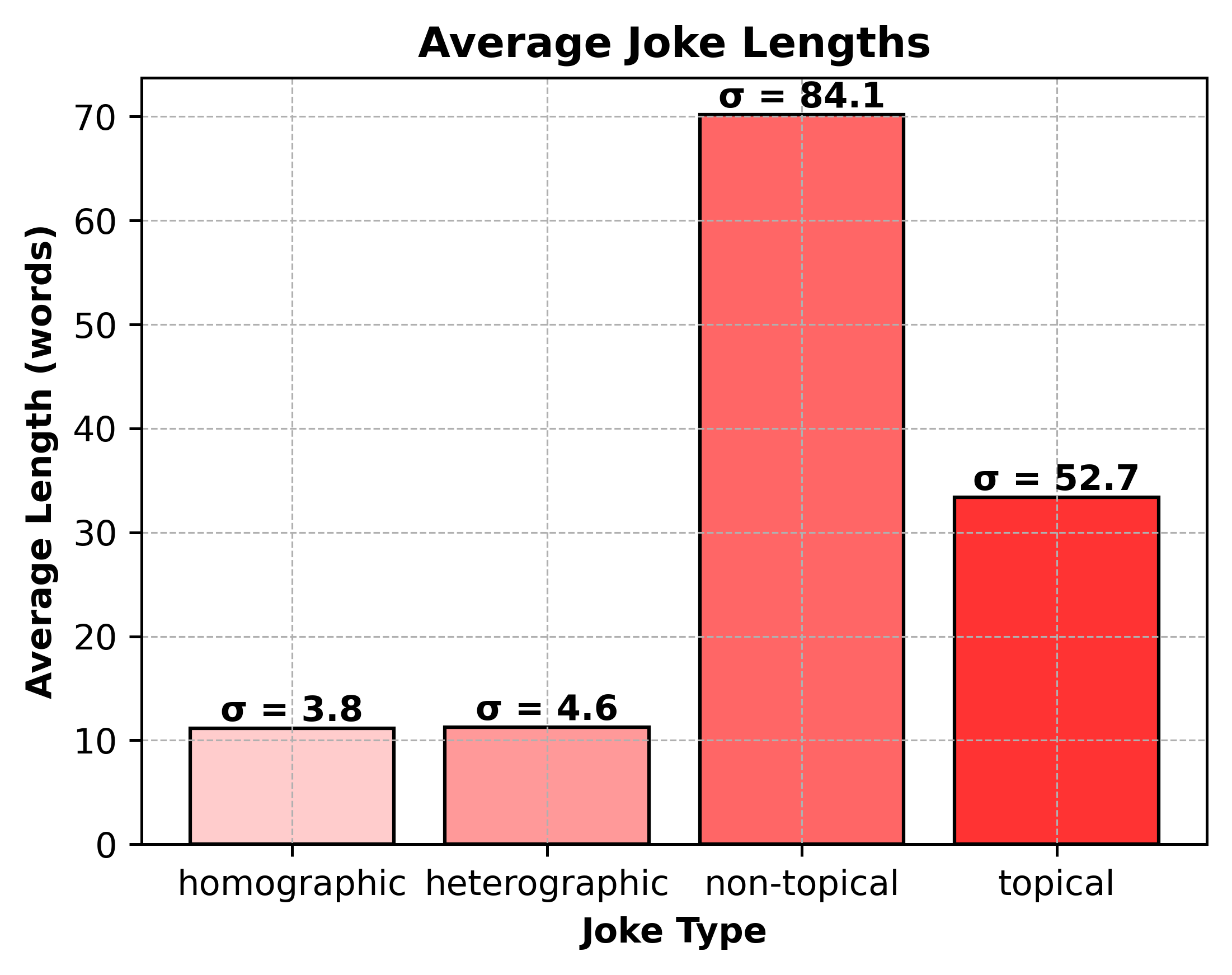}
    \caption{Average joke lengths by type (with standard deviations presented on top of each bar).}
    \label{fig:lengths}
\end{figure}

\subsection{Existing Datasets}
Many existing datasets focus on size, rather than annotation depth and consistency. For example, r/Jokes \cite{weller-seppi-2020-rjokes} presents 550k jokes taken from Reddit with metadata such as upvote count, whilst SemEval 2017 Task 7 \cite{miller-etal-2017-semeval} consists of 3387 simple puns. Whilst \citet{miller-etal-2017-semeval} add annotation of the puns' meanings via WordNet senses, there are no natural language explanations. \citet{sun-etal-2022-expunations} improve upon the SemEval dataset by crowdsourcing textual explanations for a subset, but the scope remains limited to only puns. Furthermore, \citet{hessel-etal-2023-androids} present a dataset of multimodal humour from The New Yorker cartoon caption contest, providing written explanations and Wikipedia links to required world knowledge. Finally, whilst \textit{ExplainTheJoke.com}\footnote{\url{https://explainthejoke.com/}} contains detailed joke explanations, the majority are simple puns and the explanations frequently lack objectivity, being highly editorialised for entertainment purposes.

\subsection{Our Dataset}
In the following section, we outline the process of creating our own dataset with a larger range of joke types and concise, objective explanations to form the backbone of our humour analysis. Overall, our dataset contains 600 jokes, consisting of 150 of each type: homographic puns, heterographic puns, non-topical Reddit humour, and topical Reddit humour. \autoref{fig:lengths} presents the average joke lengths within each subset, whilst \autoref{fig:examples} presents examples of each subtype (see Appendix \ref{apx:additional} for further examples).

\subsection{Topical Online Humour}
To identify jokes that require knowledge of real-world entities and events, we use a subset of \textit{r/Jokes} \cite{weller-seppi-2020-rjokes}. Whilst Reddit contains potentially unsavoury material, the presence of strong opinions on topics such as politics and pop culture phenomena results in a rich array of topical humour. Whilst r/Jokes contains a plethora of different topics, not all joke instances rely on contemporary world knowledge. We therefore perform filtering based on upvotes and the presence of named entities (see Appendix \ref{apx:implementation} for more details). Of the filtered jokes, we take the top 150 with the most upvotes. We removed jokes that did not require external knowledge, in addition to meta-level humour that relied on knowing that the joke came from Reddit (where this is not explicitly mentioned in the joke itself).

\subsection{Non-topical Online Humour}
We additionally desire to compare the results of topical humour explanation with other common forms of internet humour that do not require esoteric world knowledge. These jokes often present different forms of humour than the simple puns that have been the focus of most existing works on automatic joke explanation. Consequently, we select another 150 top-scoring jokes from the r/Jokes dataset that do not require knowledge beyond what may be considered "common sense" (e.g., cultural norms). 
Both the \textit{topical} and \textit{non-topical} subsets are checked by another linguist, who is not an author of this work, to ensure their correct categorisation. In effect, the 300 jokes that form the \textit{topical} and \textit{non-topical} categories are the top 300 highest-scoring jokes based on upvote-downvote ratio, categorised by their content.

\subsection{Traditional Puns}
We source an additional 300 simple pun-based jokes from SemEval 2017 Task 7 \cite{miller-etal-2017-semeval}, which has acted as the source data for many works on computational humour. Specifically, we take the first 150 jokes from both the homographic and heterographic training data for Subtask 3 as the basis of the \textit{homographic} and \textit{heterographic} joke subsets.

The jokes contained within the \textit{topical} and \textit{non-topical} subsets from Reddit may rely on the features of heterographic or homographic puns. However, the interpretation of these puns requires additional real-world knowledge besides the meaning of the pun words. For example, "\textit{New Teslas don't come with a new car smell. They come with an Elon Musk.}", requires knowledge regarding Elon Musk's relationship to Tesla.

\subsection{Ground Truth Joke Explanations}
As a central contribution, we write ground-truth explanations for all jokes. Due to being written by an author of this work, we perform quality control using 3 native English-speaking computer science students. Each evaluator assessed 200 joke explanations (50 per type), each seeing different jokes.\footnote{Participants were given a 30 GBP Amazon voucher for approximately 1.5 hours of evaluation.} We evaluate these explanations using the same evaluation criteria used for human evaluation in \S\ref{sec:human-eval} (with the rubric presented in Appendix~\ref{apx:rubric}), using a 0-5 Likert scale on the criteria of explanation accuracy and completeness. All evaluators scored all human-authored explanations a maximum score of 5 on both criteria or commented on required improvements, which were implemented. Minor explanation post-editing was required for less than 1\% of explanations. Further details are in Appendix~\ref{apx:implementation}.

\subsection{Additional Dataset Features}
We also present URLs to webpages containing the relevant knowledge required to understand jokes from the \textit{topical} subset, including Wikipedia pages, news articles, and blog posts. We include phonetic transcriptions of all jokes after identifying model hallucinations stemming from a lack of phonetic knowledge (details in Appendix \ref{apx:dataset-enhancement}). Neither of these fields are used in the following analyses.

\section{Methodology}
In the following section, we present our hypotheses relating to the research questions posed in \S\ref{sec:intro}, in addition to the models that we evaluate and our scoring criteria for the explanations.

\subsection{Hypotheses}
To answer the research questions presented in the introduction, we pose the following hypotheses:
\begin{itemize} [noitemsep,topsep=2pt,parsep=0pt,partopsep=0pt]
    \item \textbf{H1}. Traditional puns (homographic and heterographic) will be easier for LLMs to explain than high-rated jokes from Reddit, owing to the former's frequent reliance on the semantics and phonetics of common words.
    \item \textbf{H2}. Heterographic puns will be more difficult for models to explain than homographic puns due to the former's reliance on phonetic similarity, which is not explicitly encoded in orthographic text.
    \item \textbf{H3}. Topical humour will be more difficult for models to explain than non-topical Reddit humour, due to the former's reliance on subtle references to contemporary pop culture and events, rather than common sense reasoning and general knowledge.
    \item \textbf{H4}. Larger model variants will perform better than smaller variants, particularly for topical humour, due to being able to store larger amounts of information in their parameters regarding specific events and individuals, whilst smaller models focus on retaining common general knowledge.
\end{itemize}

\subsection{Models}
To assess the ability of state-of-the-art LLMs in this task, we select the following suite of open- and closed-source models: \textbf{Llama 3.1} (8B and 70B) \cite{llama3}, \textbf{Gemini 1.5} (Pro and Flash) \cite{gemini_15}, \textbf{GPT-4o} (standard and Mini) \cite{gpt4report}. These models represent a range of sizes, using both the flagship large variants and their lightweight counterparts, allowing the investigation of model size on performance in this task. We additionally use variants of \textbf{DeepSeek-R1} to identify the abilities of reasoning-specific models. Specifically, we use the distilled 8B and 70B variants that are based upon Llama 3.1 and 3.3, respectively.

Importantly, all of the selected models were trained on data pertaining to events up to at least 2023, with all jokes in the \textit{non-topical} and \textit{topical} subsets being posted to Reddit between 2008 and 2019. Therefore, we are fairly assessing the ability of these models to explain jokes that reference events contained within their training data, investigating the models' ability to correctly associate this knowledge with the abstract references made within topical jokes.

All models were tasked with producing an explanation around 100 words (based on the lengths of the gold standard explanations). Additional details, including the full prompt and specific models used, are presented in Appendix \ref{apx:implementation}.

\subsection{Evaluation Criteria}

To assess the quality of joke explanations generated by the models, we propose a scoring rubric with two core criteria, accuracy and completeness, each evaluated on a 6-point scale (0-5). Focusing on such criteria allows the application of the scoring rubric to all assessed joke types (such as real-world references in the \textit{topical} subset constituting completeness, or pun word definitions in traditional puns). The rubric is presented below:
\\
\label{apx:rubric}
\begin{enumerate} [noitemsep,topsep=2pt,parsep=0pt,partopsep=0pt]
    \item \textbf{Explanation Accuracy (0-5 points)}
    \begin{itemize} [leftmargin=*]
        \item \textbf{5 points}: The explanation is fully accurate, correctly identifying the joke’s main humour and context without errors.
        \item \textbf{4 points}: The explanation is mostly accurate with minor errors or omissions that do not significantly impact understanding.
        \item \textbf{3 points}: The explanation contains mostly correct elements but includes noticeable errors that may hinder understanding.
        \item \textbf{2 points}: The explanation is largely inaccurate, with only a few correct points amidst significant errors.
        \item \textbf{1 point}: The explanation is mostly incorrect, with minimal recognition of the joke’s intent.
        \item \textbf{0 points}: The explanation is entirely incorrect, failing to identify the humour or context correctly.
    \end{itemize}

    \item \textbf{Explanation Completeness (0-5 points)}
    \begin{itemize}[leftmargin=*]
        \item \textbf{5 points}: The explanation is comprehensive, covering all relevant aspects and nuances.
        \item \textbf{4 points}: The explanation covers most aspects but misses minor details.
        \item \textbf{3 points}: The explanation addresses the central humorous element but omits some important details that maximise understanding.
        \item \textbf{2 points}: The explanation is incomplete, addressing only basic aspects while missing significant details.
        \item \textbf{1 point}: The explanation is minimal, covering only a small part of the joke.
        \item \textbf{0 points}: The explanation is entirely superficial or irrelevant, failing to address any key components.
    \end{itemize}
\end{enumerate}

\textit{Accuracy} entails whether or not the explanation contains correct material, and therefore does not negatively score content where parts of the joke have not been explained, but rather assesses the presence or absence of hallucinations and misunderstandings in a response. On the other hand, \textit{Completeness} covers whether the fundamental elements of a joke have been addressed. Consequently, a joke may be fully explained (i.e., Completeness = 5), but additional information has been included which is incorrect (Accuracy $\approx$ 3). Conversely, an explanation may identify that a joke is a pun based on a specific word (i.e., Accuracy = 5), but omit explanations of the pun's meanings or other required information for understanding (i.e., Completeness $\approx$ 2). 

Due to the significant amount of time required for rating 4800 explanations ($\approx$ 40 hours), 1 author (who is a native speaker of English and holds degrees in Linguistics) rated all explanations. However, a subset of 320 explanations (10 jokes * 4 joke types * 8 models) was re-annotated by 2 third-party computer science students (who are also native English speakers).\footnote{Participants were given a 60 GBP Amazon voucher for approximately 3 hours of evaluation.} These annotations were used exclusively to verify the primary evaluator's reliability. The results showed moderate levels of agreement for both accuracy ($\alpha$ = .574) and completeness ($\alpha$ = .553) using Krippendorff's alpha \cite{Krippendorff2019}, as well as high positive Pearson's correlations \cite{PearsonsCorrelation1895} between the scores assigned by the primary human evaluator and third-party evaluators for accuracy (r = .765) and completeness (r = .795), both with $p \textless .001$. Overall, 82\% of the Accuracy scores and 84\% of the Completeness scores from the primary evaluator and the 2 third-party evaluators differed by 0 or 1 points, resulting in an 80.5\% agreement between the main evaluator and third-party evaluators when converted to a binary pass/fail (as in $\S$\ref{sec:success}). Across the whole dataset, we additionally leverage the LLM-as-a-judge paradigm using Qwen2.5-72B-Instruct, finding agreement for accuracy ($\alpha$ = .565) and completeness ($\alpha$ = .519) to be substantial, additionally supported by high Pearson's correlations for accuracy (r = .641) and completeness (r = .602), both with $p \textless .001$. See Appendix~\ref{apx:jury} for details on the LLM-as-a-judge set up, and Appendix~\ref{apx:third-party} for a deeper analysis of the ratings from the third-party evaluators and LLM judge.

\section{Human Evaluation}
\label{sec:human-eval}

 \begin{figure*}[!ht]
    \centering
    \includegraphics[width=1\linewidth]{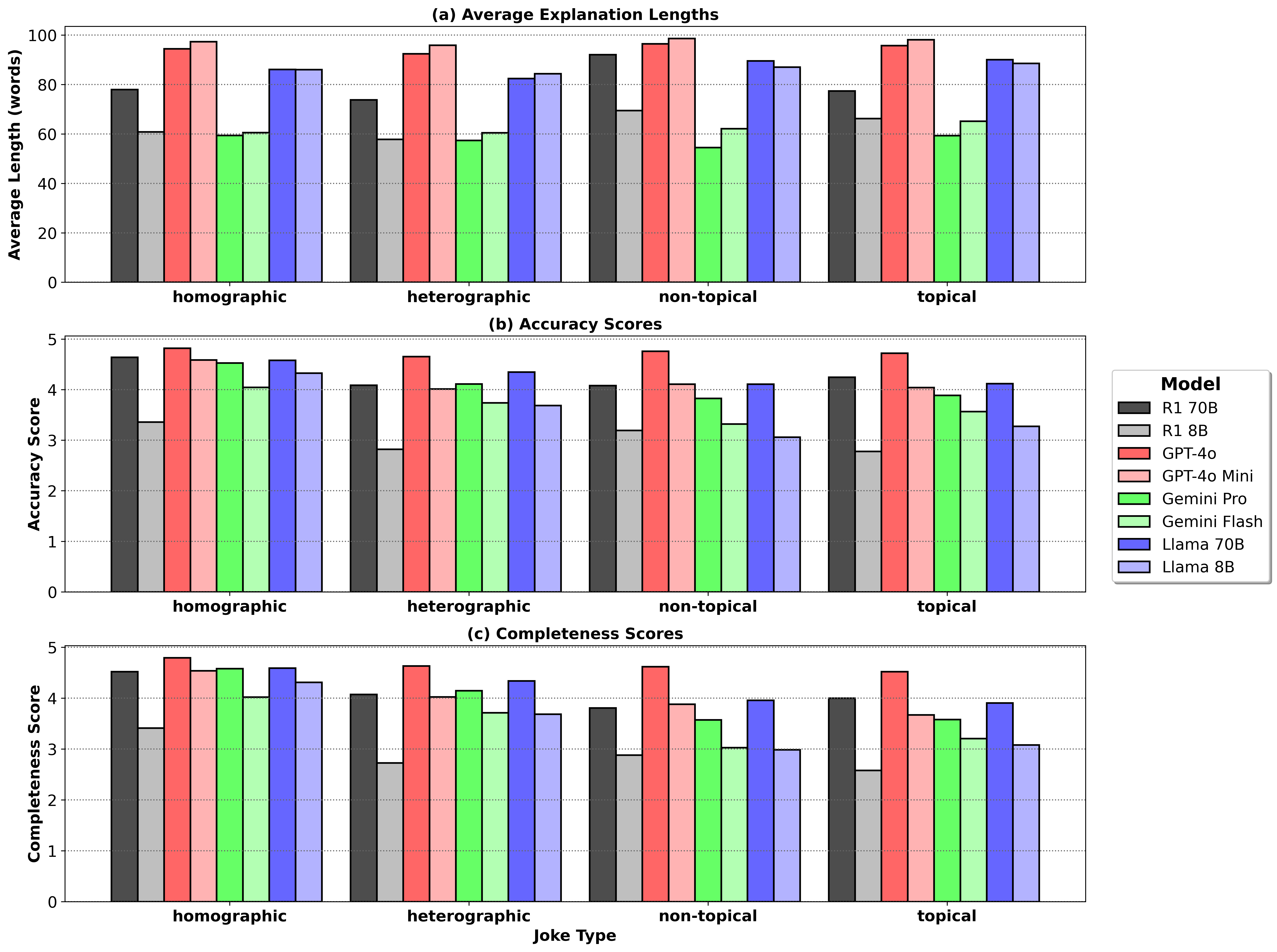}
    \caption{(a) Average explanation lengths, (b) Average Accuracy score, (c) Average Completeness score, by model and joke type.}
    \label{fig:barcharts}
\end{figure*}

Average explanation lengths alongside accuracy and completeness scores can be seen in \autoref{fig:barcharts}.
Firstly, the lengths of explanations remain consistent within models regardless of joke type, suggesting that the length of a joke does not necessarily impact the required explanation length. Notably, the Gemini-based models produce the shortest explanations across all joke types, whilst GPT-4o-based models consistently generate the longest explanations, opting for more verbose outputs. However, the shorter explanations produced by the Gemini models do not seem to negatively impact their performance in terms of accuracy or completeness. Despite being more concise, the Gemini models appear to be efficient, focusing on the essential elements of an explanation without adding superfluous detail.

In terms of overall performance, completeness scores are generally lower than accuracy scores across all models. This difference indicates that the models are more likely to miss key details in their explanations than to hallucinate incorrect information or misinterpret jokes entirely. This trend suggests that whilst the LLMs may recognise the core components of a joke, they often fail to fully elaborate on all relevant aspects that contribute to humour.

When comparing model performance, GPT-4o consistently outperforms all others, demonstrating the highest accuracy and completeness scores across joke types. Interestingly, GPT-4o Mini outperforms Gemini Pro in some specific cases, such as in accuracy on homographic jokes, as well as in both accuracy and completeness for non-topical and topical jokes. Additionally, open-source Llama 70B, outperforms most other models except GPT-4o, whilst the R1-based reasoning models underperform compared to most other non-reasoning models.

\subsection{Explanation Success Rate}
\label{sec:success}

 \begin{figure*}
    \centering
    \includegraphics[width=1\linewidth]{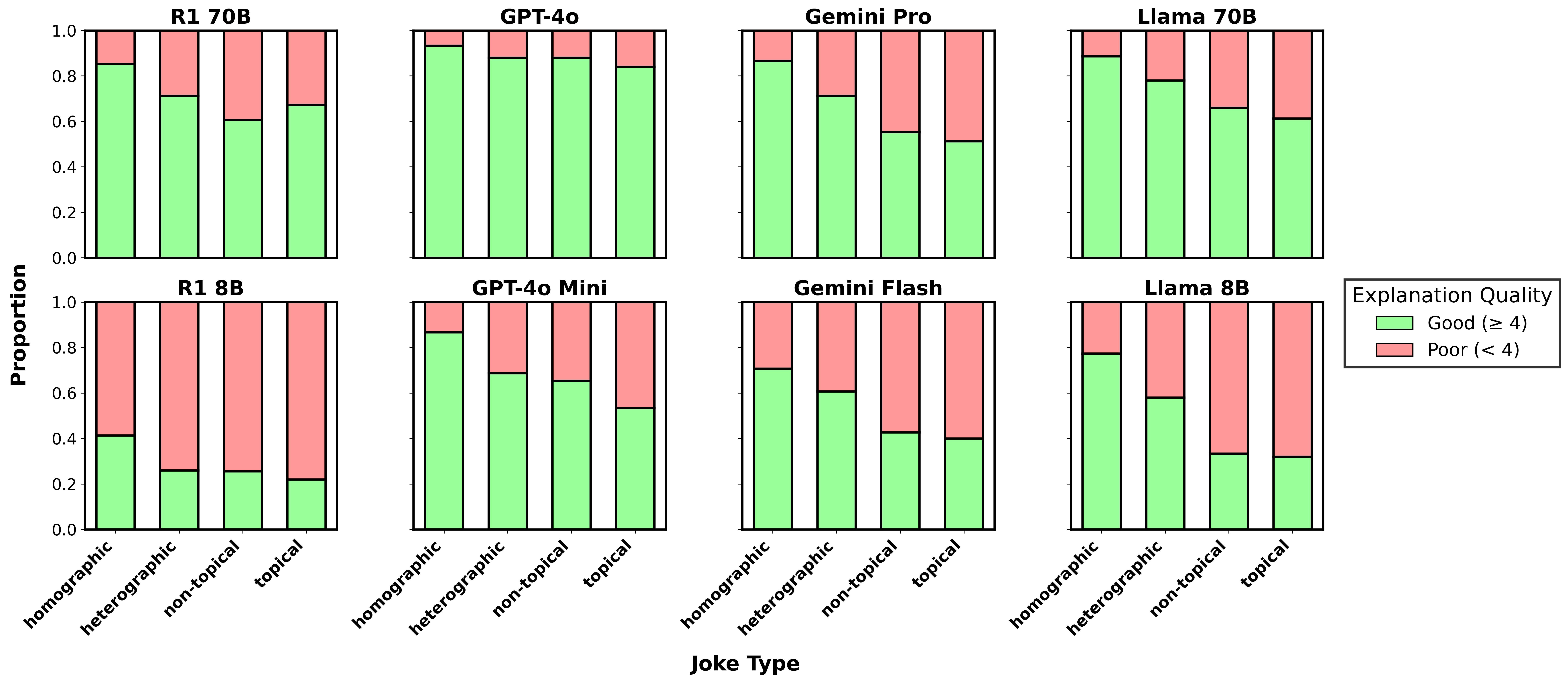}
    \caption{Proportion of "good" and "poor" joke explanations by model and joke type. A "good" explanation is defined as one that scores at least 4 on both Accuracy and Completeness.}
    \label{fig:binary}
\end{figure*}

For an explanation to be useful, it must be both accurate and complete, rather than scoring highly on only one criterion. As a result, in \autoref{fig:binary} we present a binary categorisation of explanations, treating scores of 4 or above on both criteria as being "\textit{good}" quality explanations (i.e., containing only minimal, non-intrusive errors, and/or omitting minor details). This facilitates a clear comparison of the challenges different joke types pose. 

As illustrated in \autoref{fig:binary}, the success rate of explanations varies significantly across different joke types. Homographic jokes, where a word has multiple meanings but identical spelling, consistently yield the highest proportion of successful explanations across nearly all models. This suggests that it is easier to navigate the ambiguity of homographs, likely due to language models being aware of the polysemy of the different pun words.
In contrast, heterographic jokes, which involve words that sound the same but are spelt differently, pose more difficulty for all models. GPT-4o is the only model that maintains a relatively high success rate in this category. The challenge likely stems from the additional complexity introduced by needing to recognise the phonetic similarity of the pun word whilst being trained on orthographic tokens.
Non-topical jokes also show varying levels of performance. The non-topical subset consists of jokes from Reddit, including puns, incongruities and situational irony that cover a wide range of joke forms that have been rated highly by the Reddit userbase, causing the LLMs to struggle more severely than traditional puns.
Lastly, topical jokes, which require contextual awareness of specific events, present the most notable challenges. While GPT-4o generates a relatively high proportion of "good" explanations, other models are substantially less successful. This is likely due to the subtle references to pop culture and real-world entities in the "topical" subset presenting a significant challenge to current state-of-the-art LLMs, requiring both high levels of reasoning in addition to a form of information retrieval where the information that needs retrieving is not always explicitly referred to. 

We see a unique pattern of performance from the Deepseek R1 models. Whilst R1 70B performs well overall, the model's reasoning capabilities do not result in it outperforming any other tested model. On the other hand, the 8B variant is the worst-performing model. We hypothesise that this is due to the focus on "reasoning" surrounding mathematical and coding problems, rather than "reasoning" in the general sense, such as identifying the common-sense incongruities that frequently cause humour, or "overthinking" the humour in simple puns. Furthermore, due to the reasoning steps taken, it may be the case that R1 8B begins with a moderately good answer (in line with the other models tested), but due to its uncertainty in explaining the jokes, the reasoning steps introduce many opportunities for hallucination, resulting in the smaller 8B model continuing to follow incorrect reasoning before presenting its final answer, therefore becoming progressively worse.

We perform a logistic regression analysis, which further supports our hypotheses. Model size has a strong and significant impact on explanation quality, with larger variants being far more likely to generate "good" explanations ($\beta$ = 1.707, p < 0.001). This reinforces our finding that larger models significantly outperform smaller ones. Regarding joke difficulty, the results align with the predicted ordering. Homographic jokes are the easiest to explain ($\beta$ = 0.583, p < 0.001), while non-topical ($\beta$ = -0.511, p < 0.001) and topical jokes ($\beta$ = -0.574, p < 0.001) are significantly harder. This confirms that jokes requiring external context pose a greater challenge to LLMs than those based on simple semantics and phonetics alone. Overall, the model is highly significant (p < 0.001) and explains a meaningful portion of the variance (pseudo R² = 0.138).

\section{Automatic Evaluation}
In addition to the previously presented human evaluation, we provide the results of automatic metrics in \autoref{tab:auto}. Whilst existing reference-based automatic metrics are not well designed for this task, we do confirm our originally hypothesised ordering of joke difficulties, with the explanations of homographic/heterographic puns scoring higher than the non-topical joke subset, which in turn scores higher than the topical jokes subset on most metrics.

\begin{table}[h]
\small
\centering
\begin{tabular}{lcccc}
\toprule
& \textbf{Het.} & \textbf{Hom.} & \textbf{Non-Topical} & \textbf{Topical} \\
\midrule
\textbf{SacreBLEU} & 8.53 & 10.15 & 7.86 & 7.09 \\
\hdashline
\textbf{ROUGE-1} & 0.41 & 0.43 & 0.41 & 0.40 \\
\textbf{ROUGE-2} & 0.12 & 0.15 & 0.13 & 0.12 \\
\textbf{ROUGE-L} & 0.25 & 0.28 & 0.25 & 0.23 \\
\hdashline
\textbf{METEOR} & 0.37 & 0.39 & 0.32 & 0.32 \\
\hdashline
\textbf{BERTScore} & 0.88 & 0.89 & 0.88 & 0.87 \\

\bottomrule
\end{tabular}
\caption{Automatic evaluation of LLM explanations by joke type, with human explanations as the reference. \textit{Het} and \textit{Hom} refer to heterographic and homographic puns, respectively. Further breakdown by model and joke type can be found in Appendix \ref{apx:full-auto}.}
\label{tab:auto}
\end{table}

\section{Case Study}

\begin{figure*}[hbt!]
    \centering
    \includegraphics[width=1\linewidth]{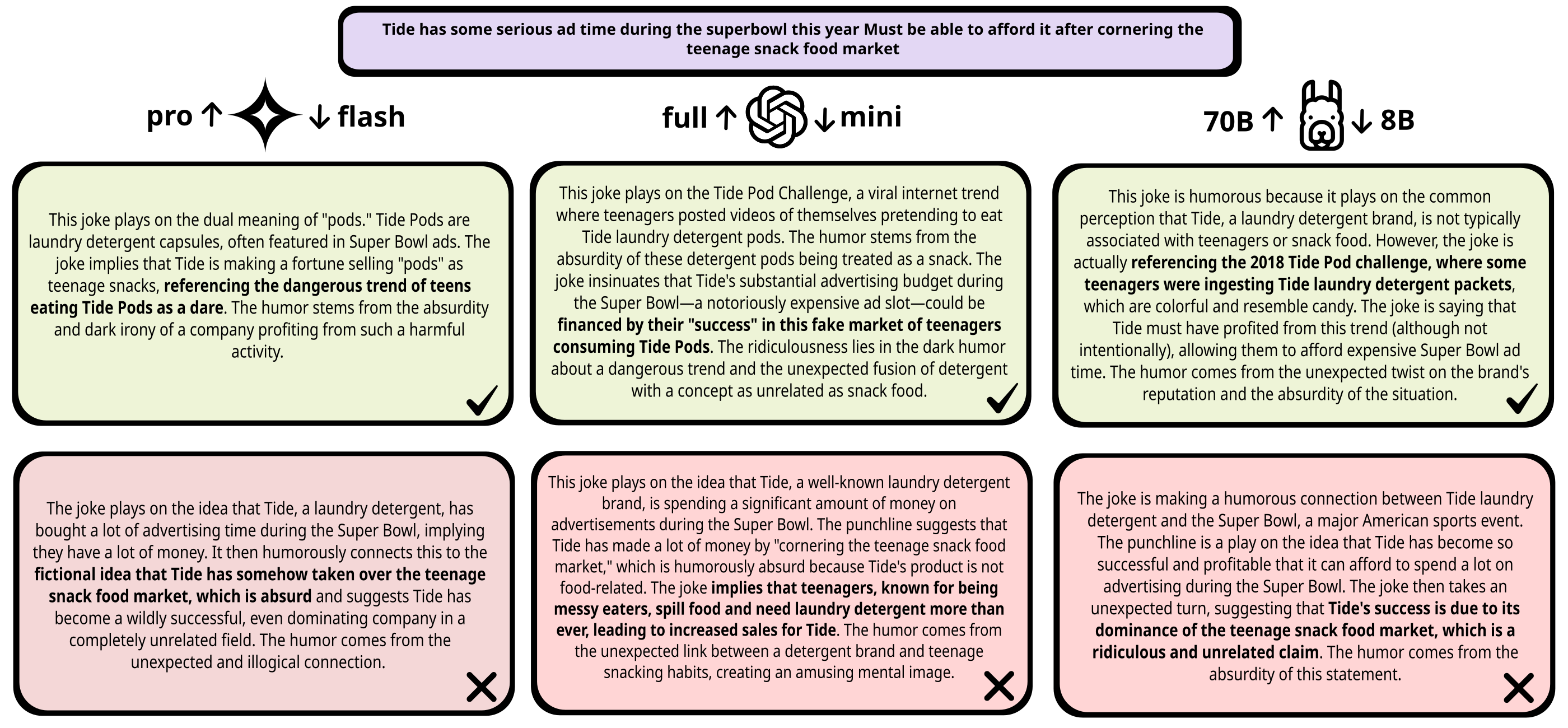}
    \caption{Case study of LLM explanations of a topical joke surrounding the "Tide Pod Challenge". From left to right: GPT-4o, Gemini 1.5, and Llama 3.1. The larger models are along the top, with the smaller variants along the bottom.}
    \label{fig:topical_case_study}
\end{figure*}

Example LLM explanations of a randomly selected joke from the \textit{topical} subset are shown in \autoref{fig:topical_case_study}. We omit the R1 reasoning models for space. The joke of interest refers to a 2018 internet phenomenon called the "Tide Pod Challenge", where teenagers would dare each other to consume toxic laundry pods (\textit{Tide} being a specific brand). The central element to the joke's explanation is the suggestion that this new "teenage snack food market" increased sales, ergo generating enough revenue for the company to afford notoriously expensive advertising at the Super Bowl American football event. Interestingly, all of the full-size models inferred the reference to this challenge in their explanations. However, the smaller counterparts consistently omitted this information or presented misinterpretations in their explanations. GPT-4o Mini, for example, explains the joke by suggesting that "teenagers are known for being messy eaters", which caused increased sales. On the other hand, Llama 8B and Gemini Flash focus on the idea that the humour stems from "absurdity", without making any explicit mention of the real-world phenomenon, thereby failing to inform the user of important context.
Further discussion of explanation characteristics can be found in Appendix \ref{apx:additional}.

\section{Discussion}
Firstly, regarding \textbf{H1}, that traditional puns will be easier for models to explain than highly-rated jokes from Reddit, we found this to be true, with the non-topical and topical subsets significantly impacting explanation success rates.

Similarly, \textbf{H2} concerned the hypothesis that homographic puns (which rely on polysemy) are easier to explain than heterographic puns, which was found to be the case in our tests. We observed that heterographic puns more frequently lead to hallucination (i.e., a lower accuracy score), owing to their reliance on phonetic characteristics, with LLMs having only implicit knowledge of how similar different words may "sound".

\textbf{H3} posited that the \textit{topical} subset would present the most difficulty for LLMs. Our testing revealed a statistically significant decrease in explanation success rates for topical humour compared to all other joke formats, with many explanations falling short of correctly identifying contemporary real-world references.

Finally, \textbf{H4} stated that the larger models would outperform smaller variants. Our results showed that this was the case, with R1 70B outperforming R1 8B, GPT-4o outperforming Mini, Gemini Pro outperforming Gemini Flash, and Llama 70B outperforming Llama 8B.

As predicted, the performance discrepancy was most noticeable within the \textit{topical} subset. Whilst these jokes were significantly shorter than the \textit{non-topical} subset (see \autoref{fig:lengths}), their reliance on subtle references to esoteric entities and events required the models to identify information that was not explicitly stated if they were to correctly identify the humour. This creates an environment ripe for hallucinations, as the models attempt to generate an explanation for a joke where they often cannot identify the location of the punchline without additional information. We further hypothesise that the larger models struggled less with topical jokes, relative to smaller models, due to the additional parameters allowing the storage of more specific knowledge that does not appear in the training data as frequently as common sense concepts (e.g., that hot things cause burns, or that sexual relations with a blood relative are taboo).

\section{Conclusion}

We presented the first empirical assessment of how well LLMs are able to explain the humour found within jokes of varying types, including simple puns and jokes that rely on contemporary topical knowledge. Our findings demonstrate that the primary focus on puns in prior work is not representative of the abilities of LLMs to accurately and comprehensively explain a more broad array of joke types. Furthermore, we also observe that no existing state-of-the-art models can consistently explain jokes of any type correctly. 

\section*{Limitations}
In this work, we investigated the extent to which joke format and complexity affect the ability of LLMs to generate accurate and complete explanations. However, the nature of topical humour is that it is ever-evolving, with jokes relating to developing news stories and pop culture events continuously appearing. Consequently, our work focuses only on a subset of possible jokes and does not assess the ability of models to explain highly recent jokes, of which there may be only limited knowledge contained within a given model's training data (as the volume of reporting and discussion on a given topic grows over time). 

Additionally, we performed a comparatively small-scale analysis of 600 jokes (150 per type). Whilst this is necessary to enable manual human assessment of the generated explanations as well as facilitate the authoring of gold-standard reference explanations, a wider range of jokes would allow deeper insight.

Due to the jokes being from secondary sources (i.e., not novel jokes written for the purpose of this work), it is likely that the majority of the jokes have been seen before by the LLMs. However, explanations relating to these jokes are much less likely to be present. For the Reddit jokes in particular, the presence of explanations in the training data is reliant on individual Reddit users asking for an explanation within the relevant thread, and someone replying. Whilst this is possible, we sorted the r/Jokes dataset based on upvotes, and intuitively it is less likely that popular jokes require explanations, with Reddit's userbase already possessing the required knowledge in most cases (hence the jokes receiving a significant amount of upvotes). We manually verified a subset of the jokes and found readily available explanations to be rare.

Whilst a substantial portion of our evaluation is performed by 1 author, we robustly prove the reliability of the evaluation through comparisons with third-party annotators and an LLM-judge. Owing to using Likert scales with a degree of inherent subjectivity, agreement statistics in the interval between 0.4 and 0.6 are considered good, whilst our general correlations among the different methods are very high.

Finally, myriad joke formats exist, and more fine-grained categorisation of joke types is possible.

\section*{Ethics Statement}
We believe in and firmly adhere to the Code of Conduct in the performance of this work and the methods involved. Whilst the dataset element of our benchmark contains potentially offensive materials (reflecting the often controversial nature of humour found on Reddit), the aim of this resource is to offer explanations of potentially esoteric jokes in order to clarify where particular parties may be perceiving humour from. In this work, we do not create novel humour, nor present the included dataset as appropriate for training humour generation models. Whilst detoxifying and debiasing language models is a growing interest in the NLP community, humans are fundamentally opinionated and biased in ways that others may take offence at. Whilst we agree language models should not perpetuate these biases, the authors of this work believe that, if queried, language models should be able to make use of knowledge pertaining to human beliefs and opinions in offering impartial, objective explanations of phenomena such as humour. We therefore do not view the creation of an \textit{explanation} for a joke to be equatable with an \textit{endorsement} of said joke, but rather as an ideologically neutral task. We additionally consciously write our humour explanations in an objective manner, not lending unjust credence to unverified scandals or presenting opinions as facts.
The study was improved by the internal ethics board of the lead author's institution, and all evaluators were provided with a particpant information sheet and consent form.

\subsubsection*{Acknowledgments}
Tyler Loakman is supported by the Centre for Doctoral Training in Speech and Language Technologies (SLT) and their Applications funded by UK Research and Innovation [grant number EP/S023062/1].

\bibliography{custom}

\appendix
\section{Appendix}
\subsection{Implementation Details}
\label{apx:implementation}
\paragraph{r/Jokes Filtering}
In order to select 150 jokes from the r/Jokes dataset to represent the \textit{topical} subset. we  sort the original dataset by the assigned "score" field (calculated by \citet{weller-seppi-2020-rjokes} based on the upvote-downvote ratio) to find the most popular jokes. Following this, we performed Named Entity Recognition (NER) via SpaCy, keeping jokes if they were found to contain entities of the following types from the SpaCy NER taxonomy: ORG (organisation), FAC (facility), WORK\_OF\_ART, LAW, ORG (organisations), PROD (products), EVENT, and PERSON. With these entity types consisting mainly of proper nouns of people, places and events, this filtering allows us to easily identify r/Jokes entries that are likely to require potentially esoteric world knowledge to perceive the humour within, as opposed to common sense or definitions of dictionary terms. This process reduced the dataset to just over 153k samples, of which we then take the top 150 highest scoring following manual verification.
\paragraph{Prompt}
For all models, we used the following prompt: "\texttt{Explain the following joke (presented in square brackets) in approximately 100 words.$\backslash$n$\backslash$n[JOKE]$\backslash$n$\backslash$n}". This is based on good practice for clearly marking the boundaries between the task and the joke instance, whilst the 100-word length is based on the average length of the human-authored explanations.
\paragraph{Models} In our work, we specifically made use of the following model variants: GPT-4o (gpt-4o-2024-05-13), GPT-4o Mini (gpt-4o-mini-2024-07-18), Gemini 1.5 Pro (gemini-1.5-pro-001), Gemini 1.5 Flash (gemini-1.5-flash-001), Llama 3.1 8B (Meta-Llama-3.1-8B-Instruct), Llama 3.1 70B (Meta-Llama-3.1-70B-Instruct), Deepseek R1 8B (DeepSeek-R1-Distill-Llama-8B), and DeepSeek R1 70B (DeepSeek-R1-Distill-Llama-8B). For GPT and Gemini, these refer to the stable releases available at the time (August 2024). All models had their settings (e.g., temperature) at default. We use the Google API with all user-accessible safety limits disabled, due to the potentially offensive content of some of the jokes.
\paragraph{Inference} OpenAI models were prompted via the OpenAI API, whilst the Google models used the Google GenerativeAI API. Llama 8B was run on a desktop PC with an RTX 4080 Super, whilst Llama 70B, both R1 models, and the LLM Judge was run on an HPC node using two A100s.

\paragraph{Participants} We recruited 2 third-party participants (i.e., not authors of this work) to perform evaluation on a subset of 240 explanations to calculate agreement. In compensation, they received a 60 GBP Amazon voucher for approximately 3 hours of work. Furthermore, 3 participants also evaluated the gold standard reference explanations (200 each), for which they received a 30 GBP Amazon voucher. Their involvement was conducted with the full ethical approval of the institution that the authors represent, with consent forms signed and a participant information sheet provided. The third-party checks of joke categorisation did not involve monetary compensation.

\subsection{Dataset Enhancement}
\label{apx:dataset-enhancement}
We additionally provide phonetic transcriptions in the International Phonetic Alphabet (IPA) of all jokes in the dataset. To do so, we utilise Anthropic's Claude Sonnet 3.5 (20241022), owing to advanced handling of factors such as abbreviations over traditional grapheme-to-phoneme (G2P) models. A subset of these transcriptions were manually verified by an author of this work who is a trained phonetician. We additionally present a test phrase of "\textit{The pikachu weighs 45 lbs and is 5ft 9". He works at the CIA but previously worked at NASA"}, in order to verify its handling of non-dictionary terms, abbreviations, initialisms and acronyms, which it did successfully. Specifically, we request transcriptions of the General American (GenAm) accent, owing to the primary userbase of Reddit with L1 English being North Americans.

When constructing the prompt, we provided a clear task description, also to Claude Sonnet, and requested that it construct its own prompt. Prior work has shown this to be an effective strategy \citep{zhouLargeLanguageModels2023}. The resulting prompt is as follows:

\textit{You are a linguistic transcription tool. For any text samples provided:$\backslash$n$\backslash$n 1. Convert EVERY sample into IPA format without exception - no samples should be skipped or partially transcribed$\backslash$n 2. Use General American (GenAm) pronunciation standards consistently$\backslash$n 3. If uncertain about any transcription, ask for clarification rather than making assumptions$\backslash$n 4. Do not proceed without transcribing ALL text samples provided$\backslash$n$\backslash$n Example input:$\backslash$n 'Hello world'$\backslash$n$\backslash$n Expected output:$\backslash$n \textipa{h@"loU w\textschwa\textrhoticity ld}$\backslash$n$\backslash$n Please proceed with the transcription."}

\subsection{Additional Annotator {\&} LLM Judge Results}
\label{apx:third-party}

\begin{figure*}
    \centering
    \includegraphics[width=1\linewidth]{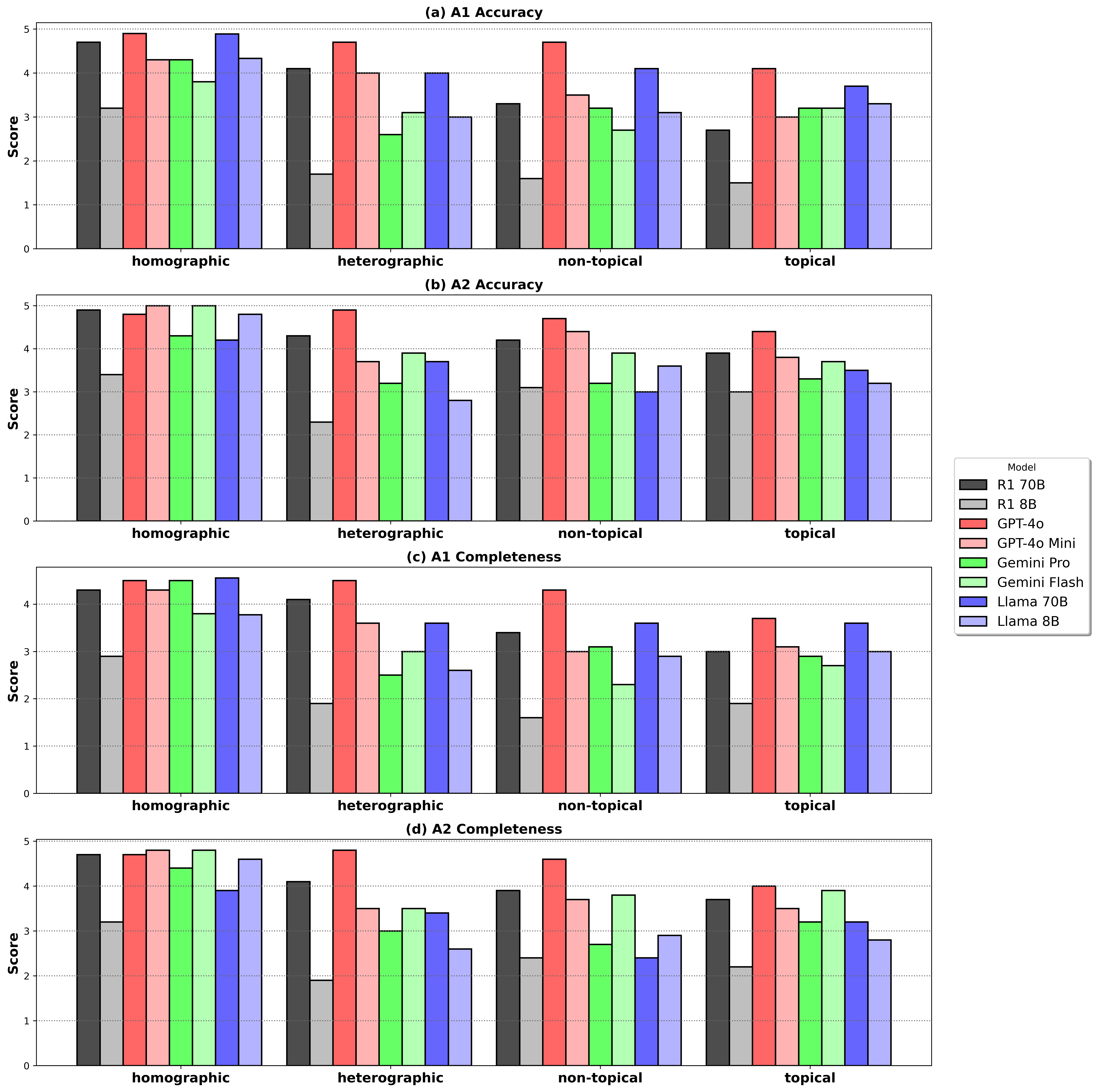}
    \caption{Human evaluation results on the metrics of accuracy and completeness on a subset of 320 explanations (10 jokes * 4 types * 8 models). A1 and A2 refer to two different annotators.}
    \label{fig:third-party-barchart}
\end{figure*}

\begin{figure*}
    \centering
    \includegraphics[width=1\linewidth]{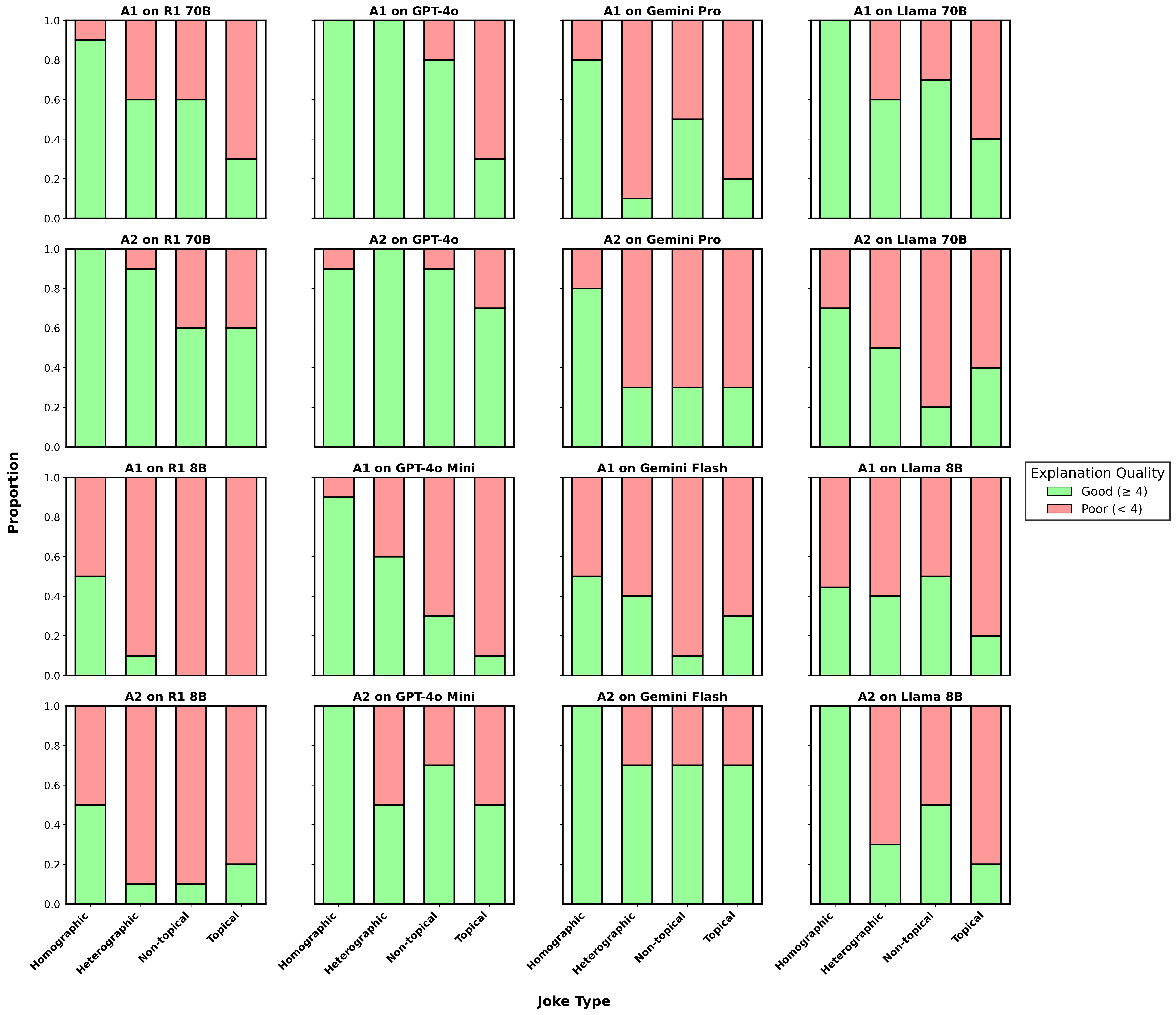}
    \caption{Binary evaluation results on a subset of 320 explanations (10 jokes * 4 types * 8 models). A1 and A2 refer to two different annotators.}
    \label{fig:third-party-binary}
\end{figure*}
    
\begin{figure*}
    \centering
    \includegraphics[width=1\linewidth]{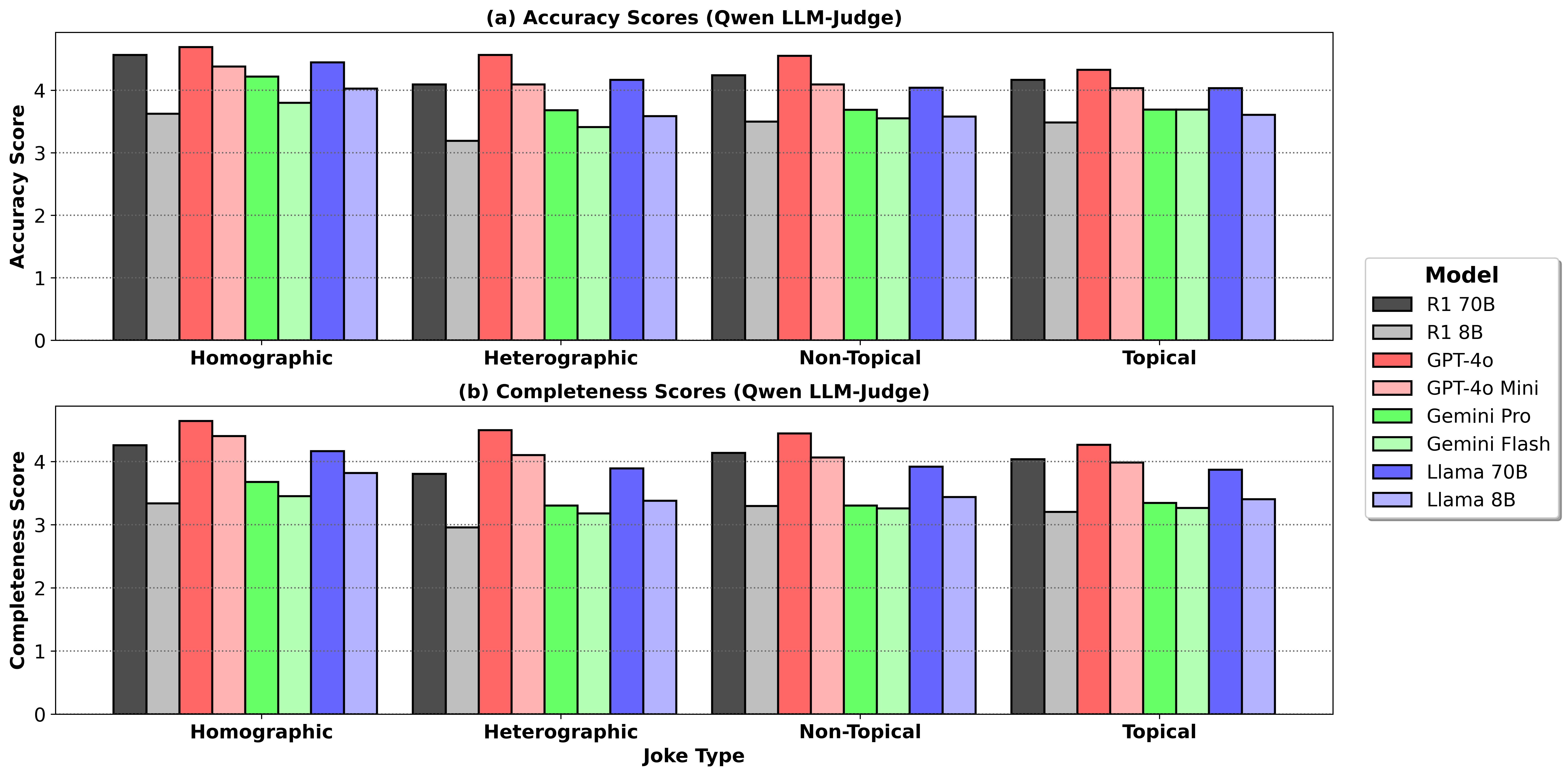}
    \caption{LLM-as-a-judge results from Qwen 72B on the metrics of Accuracy and Completeness across model and joke type.}
    \label{fig:llm-barcharts}
\end{figure*}

\begin{figure*}
    \centering
    \includegraphics[width=1\linewidth]{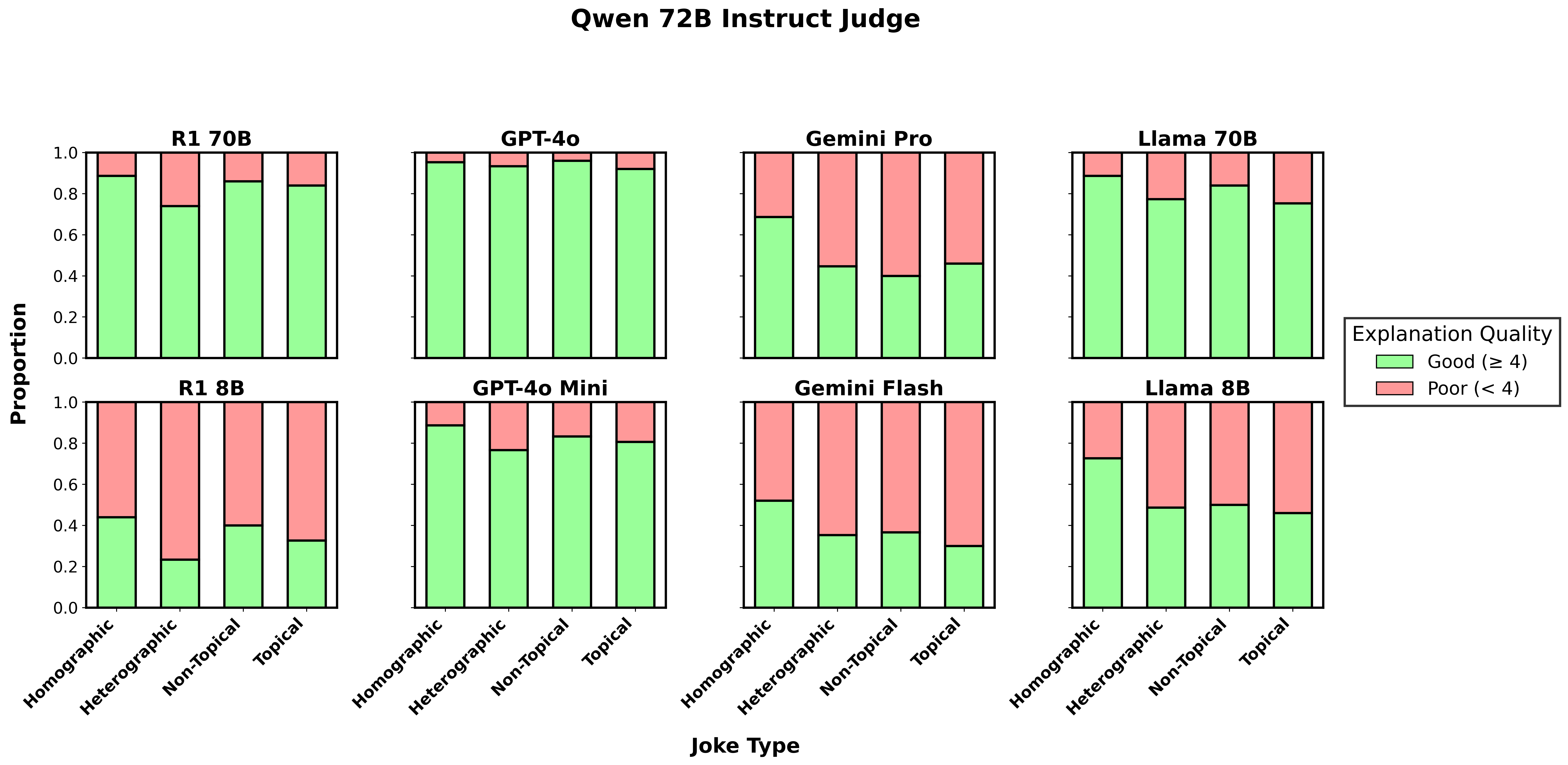}
    \caption{Binary explanation success results from the LLM-judge of Qwen 72B across model and joke type.}
    \label{fig:llm-binary}
\end{figure*}

\autoref{fig:third-party-barchart} and \autoref{fig:third-party-binary} present the results of the evaluation from the 2 third-party evaluators we recruit to analyse a subset of joke explanations. Overall, the same trend that was viewed in \autoref{fig:barcharts} and \autoref{fig:binary} can be seen, with the order of joke difficulty being the same for these evaluators on the tested subset. Owing to the smaller subsets (10 jokes per type), some minor differences can be seen in the ordering of performance for some model/joke type combinations. The same is provided in \autoref{fig:llm-barcharts} and \autoref{fig:llm-binary} for our use of Qwen 72B Instruct as an LLM judge.

\subsection{Additional Automatic Evaluation}
\label{apx:full-auto}
Full results of automatic evaluation, broken down by both model and joke type, can be seen in \autoref{tab:full-auto}.

\begin{table*}[h]
\small
\centering
\begin{tabular}{llcccc}
\toprule
\textbf{Model} & \textbf{Metric} & \textbf{Hom.} & \textbf{Het.} & \textbf{Non-Topical} & \textbf{Topical} \\
\midrule
\multirow[t]{6}{*}{\textbf{GPT-4o}} & SacreBLEU & 8.51 & 10.15 & 7.85 & 7.09 \\
\cdashline{2-6}
 & ROUGE-1 & 0.41 & 0.43 & 0.41 & 0.40 \\
 & ROUGE-2 & 0.12 & 0.15 & 0.13 & 0.12 \\
 & ROUGE-L & 0.25 & 0.28 & 0.25 & 0.23 \\
 \cdashline{2-6}
 & METEOR & 0.37 & 0.39 & 0.32 & 0.32 \\
 \cdashline{2-6}
 & BERTScore & 0.88 & 0.89 & 0.88 & 0.87 \\
\cline{1-6}
\multirow[t]{6}{*}{\textbf{GPT-4o Mini}} & SacreBLEU & 7.06 & 9.27 & 6.12 & 5.30 \\
\cdashline{2-6}
 & ROUGE-1 & 0.38 & 0.42 & 0.38 & 0.37 \\
 & ROUGE-2 & 0.10 & 0.14 & 0.11 & 0.10 \\
 & ROUGE-L & 0.23 & 0.27 & 0.23 & 0.22 \\
 \cdashline{2-6}
 & METEOR & 0.33 & 0.37 & 0.30 & 0.30 \\
 \cdashline{2-6}
 & BERTScore & 0.88 & 0.89 & 0.87 & 0.87 \\
\cline{1-6}
\multirow[t]{6}{*}{\textbf{Gemini Pro}} & SacreBLEU & 7.02 & 8.15 & 4.71 & 4.54 \\
\cdashline{2-6}
 & ROUGE-1 & 0.40 & 0.41 & 0.33 & 0.35 \\
 & ROUGE-2 & 0.10 & 0.13 & 0.09 & 0.09 \\
 & ROUGE-L & 0.25 & 0.28 & 0.22 & 0.21 \\
 \cdashline{2-6}
 & METEOR & 0.30& 0.32 & 0.22 & 0.24 \\
 \cdashline{2-6}
 & BERTScore & 0.88 & 0.88 & 0.87 & 0.87 \\
\cline{1-6}
\multirow[t]{6}{*}{\textbf{Llama 70B}} & SacreBLEU & 9.13 & 10.60 & 7.95 & 7.27 \\
\cdashline{2-6}
 & ROUGE-1 & 0.42 & 0.43 & 0.41 & 0.39 \\
 & ROUGE-2 & 0.13 & 0.14 & 0.14 & 0.12 \\
 & ROUGE-L & 0.25 & 0.27 & 0.26 & 0.23 \\
 \cdashline{2-6}
 & METEOR & 0.36 & 0.37 & 0.31 & 0.30 \\
 \cdashline{2-6}
 & BERTScore & 0.88 & 0.89 & 0.88 & 0.87 \\
\cline{1-6}
\multirow[t]{6}{*}{\textbf{Llama 8B}} & SacreBLEU & 8.41 & 9.76 & 7.16 & 5.83 \\
\cdashline{2-6}
 & ROUGE-1 & 0.39 & 0.42 & 0.40 & 0.36 \\
 & ROUGE-2 & 0.12 & 0.14 & 0.12 & 0.10 \\
 & ROUGE-L & 0.24 & 0.27 & 0.24 & 0.22 \\
 \cdashline{2-6}
 & METEOR & 0.34 & 0.36 & 0.30 & 0.28 \\
 \cdashline{2-6}
 & BERTScore & 0.88 & 0.89 & 0.87 & 0.87 \\
\cline{1-6}
\multirow[t]{6}{*}{\textbf{R1 70B}} & SacreBLEU & 7.73 & 8.47 & 6.27 & 5.89 \\
\cdashline{2-6}
 & ROUGE-1 & 0.39 & 0.41 & 0.37 & 0.36 \\
 & ROUGE-2 & 0.10 & 0.12 & 0.11 & 0.10 \\
 & ROUGE-L & 0.22 & 0.24 & 0.23 & 0.21 \\
 \cdashline{2-6}
 & METEOR & 0.33 & 0.34 & 0.28 & 0.27 \\
 \cdashline{2-6}
 & BERTScore & 0.88 & 0.88 & 0.87 & 0.87 \\
\cline{1-6}
\multirow[t]{6}{*}{\textbf{R1 8B}} & SacreBLEU & 5.25 & 6.48 & 4.85 & 4.12 \\
\cdashline{2-6}
 & ROUGE-1 & 0.35 & 0.37 & 0.36 & 0.33 \\
 & ROUGE-2 & 0.08 & 0.10 & 0.09 & 0.08 \\
 & ROUGE-L & 0.21 & 0.23 & 0.22 & 0.20 \\
 \cdashline{2-6}
 & METEOR & 0.25 & 0.28 & 0.24 & 0.23 \\
 \cdashline{2-6}
 & BERTScore & 0.87 & 0.88 & 0.87 & 0.86 \\
\bottomrule
\end{tabular}
\caption{Automatic evaluation results for each tested model, broken down by joke type. We use human-authored explanations as the ground truth reference.}
\label{tab:full-auto}
\end{table*}

\subsection{Additional Qualitative Analysis}
\label{apx:additional}

\paragraph{Reddit Joke Types}

\begin{table*}
\centering
\small
\begin{tabular}{cl} 
\toprule
\multicolumn{1}{c}{\textbf{}} & \multicolumn{1}{c}{\textbf{Non-topical Joke Examples}} \\
\midrule
\textbf{1} & "What do you call a book club that's been stuck on one book for years? Church." \\
\midrule
\textbf{2} & "TIL unvaccinated children are less likely to be autistic. Because they are more likely to be dead." \\
\midrule
\textbf{3} & "Set your wifi password to 2444666668888888. So when someone asks tell them it's 12345678." \\
\midrule
\textbf{4} & "My girlfriend is like the square root of -100. A solid 10, but also imaginary."\\
\bottomrule
\end{tabular}
\caption{\label{tab:nontopical_examples}
Examples of jokes from the non-topical subset with different types of common sense knowledge.
}
\end{table*}

\begin{table*}
\centering
\small
\begin{tabular}{cl} 
\toprule
\multicolumn{1}{c}{\textbf{}} & \multicolumn{1}{c}{\textbf{Topical Joke Examples}} \\
\midrule
\textbf{1} & "If America is storming Area 51 then the Europeans can storm the Vatican. We’ll take the aliens, you get the predators." \\
\midrule
\textbf{2} & "For anyone attending Stan Lee's funeral... Make sure you stay after the ceremony is finished." \\
\midrule
\textbf{3} & "R Kelly is really changing the rap game. He takes the art out of rap artist." \\
\midrule
\textbf{4} & "Obama smoked weed growing up, and now look where he is today. Unemployed with two kids and recently evicted."\\
\bottomrule
\end{tabular}
\caption{\label{tab:topical_examples}
Examples of jokes from the topical subset with different types of assumed world knowledge.
}
\end{table*}

\autoref{tab:nontopical_examples} and \autoref{tab:topical_examples} present examples of non-topical and topical jokes from the Reddit subsets, respectively. Regarding the non-topical jokes, the first example plays on common knowledge regarding religious practices and the focus of churches on the bible. Joke 2 is a form of deadpan humour that refers to the higher mortality rates of children who are not vaccinated against preventable diseases. The third joke presents phonetic wordplay, requiring the reader to interpret "12345678" as "one 2, three 4..." and so on. Finally, joke 4 plays on the mathematical concept of imaginary numbers. On the other hand, regarding the topical jokes, joke 1 makes reference to the media franchise of Alien v. Predator, whilst additionally referring to the 2019 news phenomenon where a Facebook event was created to convince people to storm Area 51 to find aliens. The second joke refers to the Marvel film franchise and their penchant for extended scenes following the credits, in addition to Stan Lee's real-life death. On the other hand, joke 3 makes light of a celebrity scandal, involving wordplay that requires the sequence "art" to be removed from "rap artist" to identify a particular scandal involving R Kelly. Finally, the fourth joke refers to Barack Obama losing his presidency to Donald Trump, and therefore being "unemployed" and "recently evicted" (from the White House).

\paragraph{Explanation Characteristics}

During the process of evaluating LLM responses, several unique response characteristics were observed. Firstly, the Google API refused to allow the Gemini models to produce explanations for some jokes, likely due to triggering content safeguards (5 explanations across 3 jokes, where in one instance the request was only blocked by Flash). Such instances were not notably any more/less offensive than other humour found within the Reddit joke subsets. In some instances, models provided textual responses regarding their refusal to address a particular joke (however, both cases only account for $\approx$1\% of all jokes).
Furthermore, in some instances, we observed that explanations contained subjective judgments as to the appropriateness of a particular joke, with the humour being deemed "unacceptable", without acknowledging the nuance and subjectivity of such a claim. Importantly, potentially offensive content was not exclusively found in the non-topical and topical Reddit joke subsets, with many of the traditional puns containing sexual innuendo and derogatory terms. Furthermore, whilst all of the models tested are capable of producing high-quality coherent text, we observed some instances of the Llama models resorting to degenerate patterns, such as being stuck in repetitive loops. Again, these account for a very small number of jokes (\textless 1\%). Regarding the heterographic pun subset of jokes, we noticed many instances of hallucination whereby the LLMs would explain that a punning word is phonetically similar to a word that shares little phonetic similarity. This is demonstrative of the challenge that heterographic puns present to LLMs, relying on phonetic characteristics of words, whilst such models are trained on orthographic text and therefore have limited phonetic knowledge.

\subsection{LLM-as-a-Judge Models and Prompts}
\label{apx:jury}
When selecting our LLM judge for automatic evaluation, we required that candidates not be from the same model family as any of those under evaluation or from the same family as each other. Our candidates included the Mistral Mixture-of-Experts \citep{jiangMixtralExperts2024} derivative, Prometheus 2 8x7B \citep{kimPrometheus2Open2024a}, which was specifically developed for LLM-based evaluation, Qwen 72B Instruct \citep{qwenQwen25TechnicalReport2025}, and Gemma 2 9B Instruct \citep{teamGemma2Improving2024}. These were selected due to being the best performing models in the Hugging Face Judge Arena\footnote{\url{https://huggingface.co/spaces/AtlaAI/judge-arena}} (with preliminary results disabled) that met our previously outlined criteria. At the time of selection (04/02/2025), Qwen was ranked 6th with 1203 ELO and Gemma 13th with 1157. The weights for the models were obtained from their respective Hugging Face model repositories. Overall, we found Qwen alone to demonstrate the best alignment with our human evaluation (versus a jury ensemble or any other individual judge).

Judge hyperparameters were all set to the model default except for temperature, which was fixed to 0.1 for all models, rather than zero, based on prior work \cite{sanhMultitaskPromptedTraining2022}.
The scoring criteria were the same as given to the human annotators, with the exception of the ordering of score categories, which were listed in ascending order rather than descending. This was done to better align with the training prompt of Prometheus. To obtain scores from each judge, we used the following prompt:

\textit{You are an expert judge evaluating explanations of jokes. You will be given a joke, a reference explanation (human-authored), and a model-generated explanation to evaluate.$\backslash$n$\backslash$n You must evaluate the model's explanation on the matter of \{criteria\} using only the following scoring criteria:$\backslash$n$\backslash$n \{scoring\_criteria\}$\backslash$n$\backslash$n The joke:$\backslash$n \{joke\}$\backslash$n$\backslash$n Reference explanation:$\backslash$n \{reference\_explanation\}$\backslash$n$\backslash$n Model explanation to evaluate:$\backslash$n \{model\_explanation\}$\backslash$n$\backslash$n Provide a score for this joke using only the scoring criteria provided. You must only respond with a single number between 0 and 5. You must produce no other output.}

\end{document}